%% file: main.tex
\documentclass[twocolumn,10pt]{article}
\usepackage[a4paper, total={7in, 10in}, left=0.6in, right=0.6in]{geometry}
\usepackage{graphicx}
\usepackage{amsmath}
\usepackage[style=numeric,sorting=none]{biblatex} 
\usepackage{comment}
\usepackage{hyperref}
\usepackage{longtable}
\usepackage{siunitx}
\usepackage{enumitem}

\title{deepTerra - AI Land Classification Made Easy}
\author{
    Andrew Keith Wilkinson \\
    \texttt{info@deepterra.tech}
}
\date{\today}

\begin{document}

\maketitle

\begin{abstract}
\input{abstract}

\end{abstract}

\input{introduction}

\input{overview}

\input{data_collection}

\input{augmentation}

\input{training}

\input{test_predict}

\input{case_studies}

\input{future_plans}

\newpage
\printbibliography 
\end{document}

%% file: abstract.tex
\textbf{deepTerra is a comprehensive platform designed to facilitate the classification of land surface features using machine learning and satellite imagery. The platform includes modules for data collection, image augmentation, training, testing, and prediction, streamlining the entire workflow for image classification tasks. This paper presents a detailed overview of the capabilities of deepTerra, shows how it has been applied to various research areas, and discusses the future directions it might take.}


%% file: introduction.tex
\section{Introduction} \label{intro}

Land classification using satellite imagery has emerged as a critical tool in diverse applications, including urban planning, environmental monitoring, and resource management \cite{abburu_satellite_2015}. However, deploying machine learning models for such tasks often involves overcoming challenges such as data collection, dataset augmentation, and model optimization. To address these needs, deepTerra offers an integrated suite of tools that simplifies the end-to-end workflow of image classification via machine learning.

deepTerra is designed to streamline the processes of satellite image acquisition, data preparation, and training of state-of-the-art convolutional neural network (CNN) architectures. By supporting efficient labeling, data augmentation, and hyperparameter tuning, deepTerra enables researchers and practitioners to develop robust models with minimal effort. The platform supports a variety of CNN architectures, such as ResNet, MobileNet, and EfficientNet, making it versatile for a wide range of classification tasks.

This paper provides an overview of the capabilities of deepTerra, detailing its modules for data collection, augmentation, training, and prediction. Through case studies like garbage detection, private pool identification, and beehive localization, the paper demonstrates the flexibility and effectiveness of the tool in addressing real-world problems.

%% file: overview.tex
\section{deepTerra Overview} \label{overview}

deepTerra is an integrated suite of tools that support the various stages that make up image classification via machine learning. These comprise the following modules:

\begin{itemize}
\item
\textbf{Data Collection}: This module provides tools to extract suitable image patches from pre-existing images or download satellite imagery from sources like Google Earth. It also includes features for labeling and organizing datasets efficiently. When geographic coordinates are available, they are automatically recorded and associated with each image patch for precise spatial referencing.
\item
\textbf{Image Augmentation}: Collecting and labeling a large dataset of images for training a reliable model can be challenging and time-consuming \cite{shorten_survey_2019}. Image augmentation offers an effective solution by expanding the dataset through the application of common geometric transformations, such as rotation, flipping, and shifting. These techniques maximize the utility of existing data, improving the model's robustness and performance.
\item
\textbf{Training}: Once a suitable labeled training dataset has been prepared, the model training process can begin. The tool supports a variety of popular CNN architecture families, including VGG, ResNet, Inception, and MobileNet. It provides sensible default hyperparameter settings while allowing users to fine-tune these as needed. Training progress is displayed through graphical summaries and detailed metrics, such as accuracy and F1 score.
\item
\textbf{Test and Predict}: After training a model, the test and predict module can be applied. The testing phase evaluates the model's performance on existing labeled data, providing an objective measure of accuracy. The prediction phase focuses on novel, unlabeled data—the ultimate goal of the preceding steps. Results are presented both in summary and detailed formats, with options to export them as CSV or JSON files. If geographic coordinates are included, Google Maps links can be generated to reference specific image patches. Advanced features allow for the creation of heatmaps \cite{chowdhury_explaining_2021-1} and interactive Google Map overlays in HTML format.
\end{itemize}

\begin{figure}[!h]
\centering
\includegraphics[width=\columnwidth]{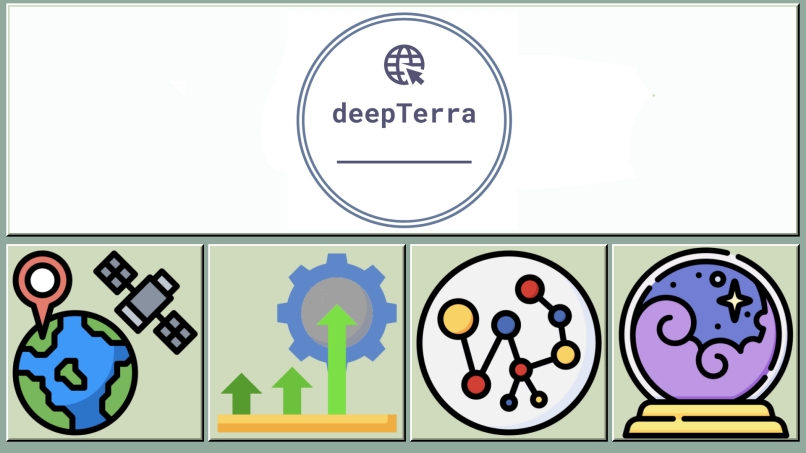}
\caption{deepTerra - home screen}
\label{fig:home}
\end{figure}

Each of the deepTerra modules are accessed through the home screen, as shown in Figure \ref{fig:home}. The following sections describe each of the deepTerra modules in detail.

%% file: data_collection.tex
\subsection{Data Collection}\label{data_collection}
\begin{figure}[!ht]
\centering
\includegraphics[width=\columnwidth]{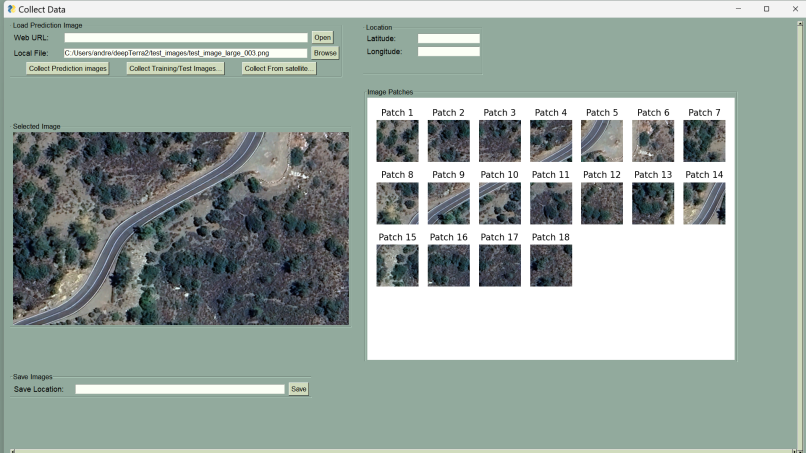}
\caption{Collecting data from a large image}
\label{fig:splitting_patches}
\end{figure}

\begin{figure}[!ht]
\centering
\includegraphics[width=\columnwidth]{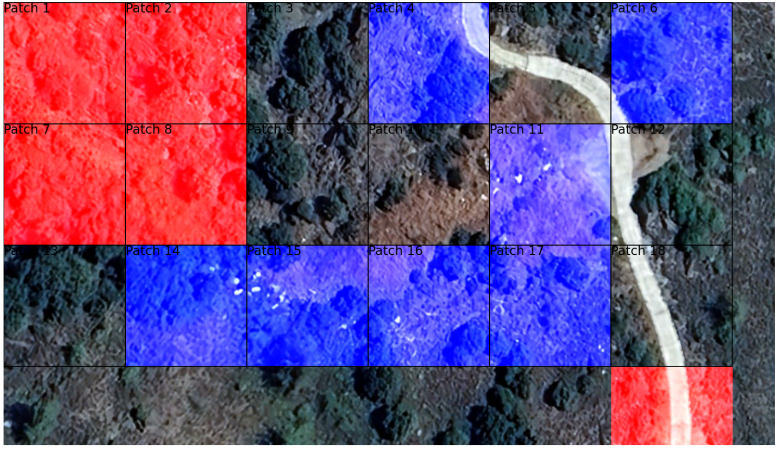}
\caption{Classifying data from image patches}
\label{fig:classifying_patches}
\end{figure}

The data collection module provides various ways to import and organize image patches.
\begin{itemize}
    \item Image patches can be created by simply loading a large existing image file, which is then automatically split into patches suitable for the machine learning process (see Figure \ref{fig:splitting_patches}). The image patches can be optionally classified by left or right clicking to indicate positive or negative.  Figure \ref{fig:classifying_patches} shows a satellite image broken into patches and a set of images specified as ``garbage" (positive, blue patches) or ``not\_garbage" (negative, red patches).
    \item Alternatively, a latitude/longitude coordinate can be supplied for the north-east corner of a \SI{1}{\square\kilo\metre} area, after which the google earth engine can be used to download and organize all the image patches for this location.  This is stored in an organized directory and consists of 1296 200x200 pixel high-resolution images.  The geographical coordinates are calculated tagged with each image patch.
\end{itemize}

The data collection module provides convenient ways to collect data which can then be used to either train models, or to perform predictions. However, these operations do not necessarily require use of the data collection tool; provided the images are stored in a suitable format, the source of the images can be produced from third party sources.

%% file: augmentation.tex
\subsection{Data Augmentation}
Collecting and classifying satellite images is often a time-consuming and challenging task. Well-curated datasets for many problem domains are difficult to find, and even when a suitable source is identified, the number of available images is often insufficient to train a robust model. Data augmentation addresses this limitation by applying various transformations to the existing dataset, effectively increasing the number of unique training samples \cite{wang_effectiveness_2017}.

deepTerra supports several standard geometric augmentation techniques, including:

\begin{itemize}[nosep]
    \item \textbf{Rotation}: Rotate each image by a random number of degrees
    \item \textbf{Shifting}: Shift an image along the vertical or horizontal axes by a random amount; effectively shifting them up, down, left, or right.
    \item \textbf{Zooming}: Scaling images up or down by a random amount.
    \item \textbf{Flipping}: Generating mirrored versions of images along the horizontal or vertical axis.
\end{itemize}

These transformations can be applied to any image dataset to expand the number of training samples while introducing variations that generate unique instances. Crucially, the augmented images retain the essential characteristics of the originals, ensuring their relevance for training. If any transformation creates ``dead" space, a configurable fill mode can be applied to seamlessly fill the gaps, maintaining the integrity of the augmented images.

\begin{figure}[!ht]
\centering
\includegraphics[width=\columnwidth]{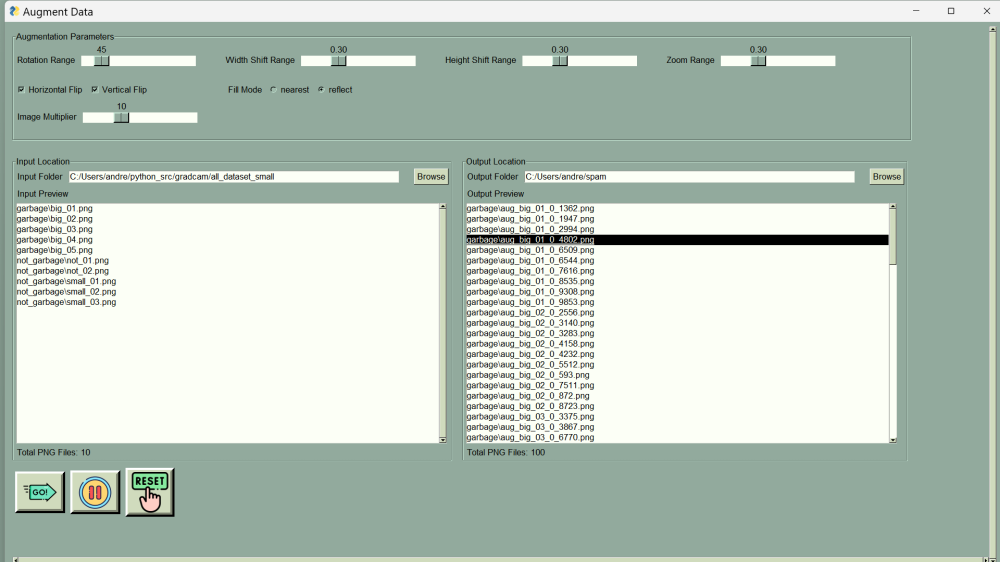}
\caption{Image Augmentation}
\label{fig:augmentation}
\end{figure}

Figure \ref{fig:augmentation} illustrates the image augmentation tool.

%% file: training.tex
\subsection{Training}
Training is the process through which a model learns to identify objects by analyzing example images. It involves utilizing a labeled dataset in conjunction with a selected machine learning architecture to process the images and extract the critical features necessary for accurately recognizing target objects \cite{sarker_machine_2021}.

deepTerra supports the training process through the following key features:

\subsubsection{Importing a dataset}
Image datasets can be imported from three main sources:
\begin{enumerate}[nosep]
    \item Local Folder: A folder on the local machine where each subdirectory corresponds to a target machine learning category (label). Each subdirectory should contain a set of images in either PNG or JPEG format.
    \item Local Compressed Archive: A .tgz archive stored on the local file system. The archive should follow the same folder and image organization as the local folder structure.
    \item Remote Compressed Archive: A .tgz archive available for download via a URL. 
\end{enumerate}

\subsubsection{Selecting a training architecture}

Training is conducted using a specific machine learning architecture. deepTerra simplifies this process by offering easy access to a diverse range of convolutional neural network (CNN) architectures, widely regarded as the ``gold standard" for AI-driven image classification tasks \cite{scott_training_2017}.

Users can simply select an architecture suitable for their task from a convenient drop-down menu. This also makes it easy to perform comparative analyses, enabling users to evaluate and identify the best-performing architecture for their specific needs.

The following CNN architectures are supported:

\begin{itemize}[nosep]
    \item ResNet50
    \item Inception V3
    \item DenseNet
    \item EfficientNet V2
    \item Inception-ResNet V2 
    \item VGG19
    \item MobileNet V3
    \item NASNet
    \item Xception
    \item ConvNet
\end{itemize}

\subsubsection{Tuning hyperparameters}
Once an architecture is selected, key parameters can be optimized to balance various aspects of the machine learning process, such as memory efficiency, training speed, and model generalization. deepTerra simplifies this process by providing sensible default settings tailored to the selected dataset and architecture. Users also have the flexibility to override these defaults with their own custom values, ensuring the system adapts to specific requirements and preferences.

The following parameters are currently supported:

\begin{itemize}[nosep]
    \item maximum number of epochs
    \item batch size
    \item early stopping
    \item dropout
    \item optimizer
    \item learning rate
    \item activation function
    \item pre-existing weights (imagenet by default)
    \item Training/validation set split
\end{itemize}

\subsubsection{Controlling the training run}
Once a dataset is loaded and an architecture selected, the training process can be initiated. As the training progresses, performance metrics are visualized through a graphical summary, complemented by a detailed text console. These outputs provide insights into each training epoch, including accuracy and loss statistics.

The training process offers flexibility, allowing users to pause, resume, stop, or reset the run at any time, ensuring complete control over the workflow.

\subsubsection{Assessing the results} \label{assessing_results}
Once a training run is completed, the results can be thoroughly reviewed. deepTerra provides a detailed summary, including a confusion matrix and key performance metrics such as accuracy, precision, recall, F1-Score, and Matthew’s Correlation Coefficient (MCC). These metrics offer a comprehensive initial assessment of the model's performance, helping users gauge its effectiveness and reliability.

\begin{figure}[!h]
\centering
\includegraphics[width=\columnwidth]{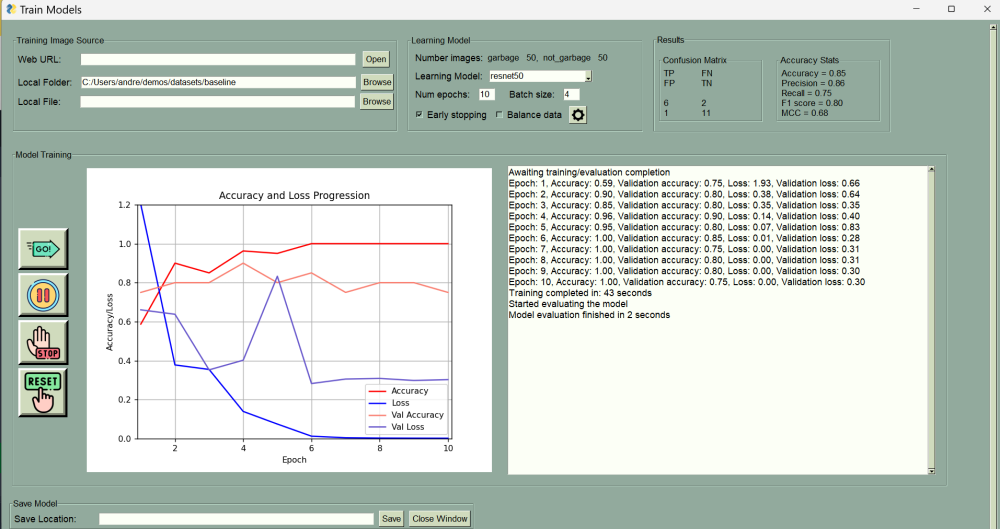}
\caption{Training a model}
\label{fig:training}
\end{figure}

Figure \ref{fig:training} illustrates the training tool.

After training is complete, the model can be used to make predictions on new, unseen image sets—fulfilling the ultimate goal of the entire process.

%% file: test_predict.tex
\subsection{Test and Prediction}
The test and prediction tool addresses two closely related scenarios:
\begin{enumerate}[nosep]
    \item Testing on Labeled Datasets: Performing predictions on test datasets to evaluate how well the model generalizes to data that differs from the training set.
    \item Predicting on Unlabeled Data: Making predictions on novel, unlabeled datasets to generate insights or classifications for new data.
\end{enumerate}

\subsubsection{Testing the Model} \label{testing}

The training process includes evaluating the model using a randomly selected subset of the training data. While this provides an initial measure of the model's accuracy, it is tied to the dataset used for training and may reflect biases present in the data collection process. To ensure the model’s ability to generalize to broader datasets, it is valuable to use an independently collected dataset for evaluation.

The testing module supports this by allowing users to load a separate dataset and select a previously trained model. During testing, predictions are generated for the dataset, with progress and sample-level results displayed in a detailed output console.

\begin{figure}[!h]
\centering
\includegraphics[width=\columnwidth]{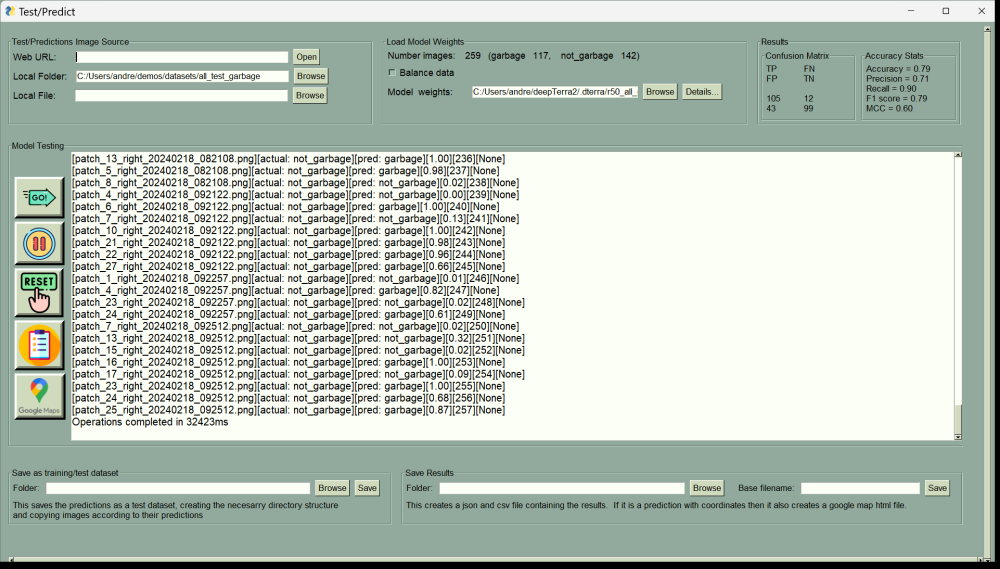}
\caption{Testing a model}
\label{fig:testing}
\end{figure}

Figure \ref{fig:testing} illustrates the testing tool.

\subsubsection{Using the Model for Predictions} \label{predictions}

The ultimate goal of training and testing a model is to deploy it for real-world predictions. While prediction also requires datasets, these are typically unlabeled—consisting solely of images without predefined classes. The data collection module can assist in gathering and organizing such datasets, just as it does for training datasets (see Section \ref{data_collection}).

Unlike testing, where performance metrics like confusion matrices can be calculated, predictions on unlabeled datasets do not allow for such evaluations because the true class of each image is unknown. However, the tool provides a summary of the prediction results, including the number of instances predicted for each class and detailed information for each prediction (see Section \ref{pred_details}).

\begin{figure}[!h]
\centering
\includegraphics[width=\columnwidth]{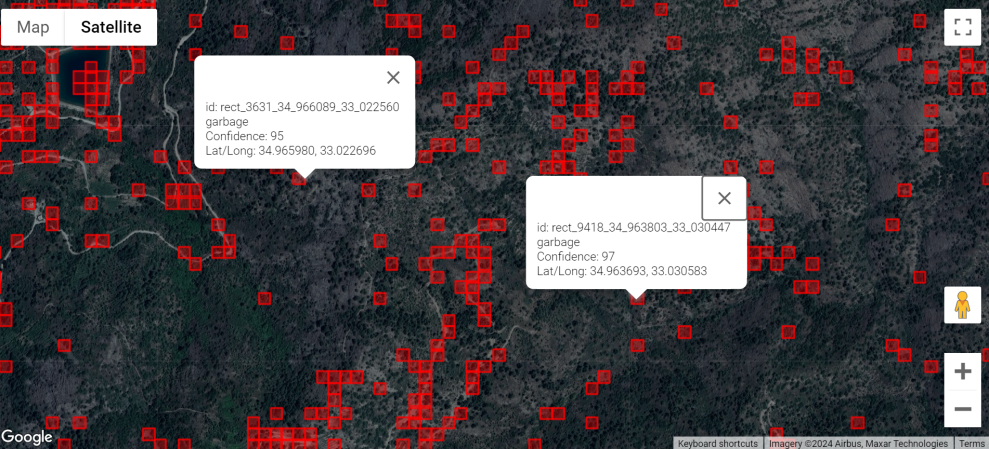}
\caption{Google maps overlay}
\label{fig:google_overlay}
\end{figure}

After completing a prediction run, the raw results are displayed in the main window, appearing similar to those from the testing variant of the tool (as shown in Figure 6). The results can also be saved in several formats for further analysis or integration into workflows:

\begin{itemize}[nosep]
\item Labeled Dataset Creation: The prediction set can be saved as a labeled dataset, with the predicted labels assigned to each image. This feature is useful for generating new training or testing datasets based on the prediction outcomes.
\item Export Options: Prediction results can be exported as a collection of files in CSV, JSON, and HTML formats. The HTML file includes a Google Maps overlay, visually representing the geographic locations of positive samples. An example of this is shown in Figure \ref{fig:google_overlay}.
\end{itemize}

In both the testing and prediction variants of the tool, a details subtool can be accessed.  This is described below.

\subsubsection{Details Window} \label{pred_details}

\begin{figure}[!h]
\centering
\includegraphics[width=\columnwidth]{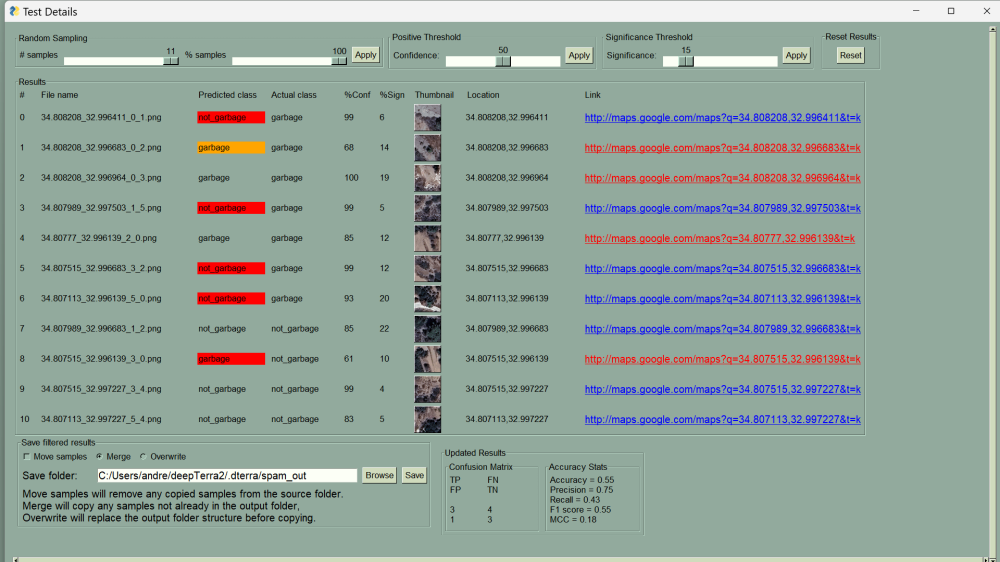}
\caption{Test/Prediction Details}
\label{fig:details}
\end{figure}

This subtool provides additional information and functionality, enhancing the utility of both testing and prediction workflows. This presents the raw test/prediction results in a more structured way, as a table of results, as shown in Figure \ref{fig:details}.  The following data points are displayed:

\begin{itemize}[nosep]
    \item filename of the image on disk or tgz archive file
    \item predicted class based on model used
    \item actual class for test runs, or chosen class for predictions where this has been defined
    \item confidence of the prediction, expressed as a percentage
    \item significance of the prediction, expressed as a percentage of pixels in the image that contribute positively to the prediction
    \item clickable thumbnail of the image (see below)
    \item location of the image expressed as a geographical latitude/longitude coordinate (if available)
    \item html link to location of the image on a google map (if available)
\end{itemize}

If present, the confusion matrix and accuracy stats are displayed.  It is also possible to save the current dataset status if it has been edited.

\subsubsection{Result Manipulations}
The results from predictions can be manipulated to focus on specific criteria, enabling more targeted analysis. Key manipulations include:

\begin{itemize}[nosep]
    \item Filtering by Confidence: Predictions can be filtered based on their confidence level—the calculated probability that a prediction is correct. For instance, users might choose to focus on predictions with confidence levels greater than 95\%.
    \item Filtering by Significance: Results can also be filtered by the percentage of an image contributing positively to the prediction. This is particularly useful for identifying predictions with large coverage areas, such as extensive garbage sites.
    \item Random Sampling: A random subset of predictions can be generated from the dataset. This feature is useful for evaluating the overall accuracy of a prediction run by reviewing a representative sample of the results.
\end{itemize}

These manipulations provide flexibility in analyzing prediction results, enabling users to tailor the outputs to their specific needs or criteria.

\begin{figure}[!h]
\centering
\includegraphics[width=\columnwidth]{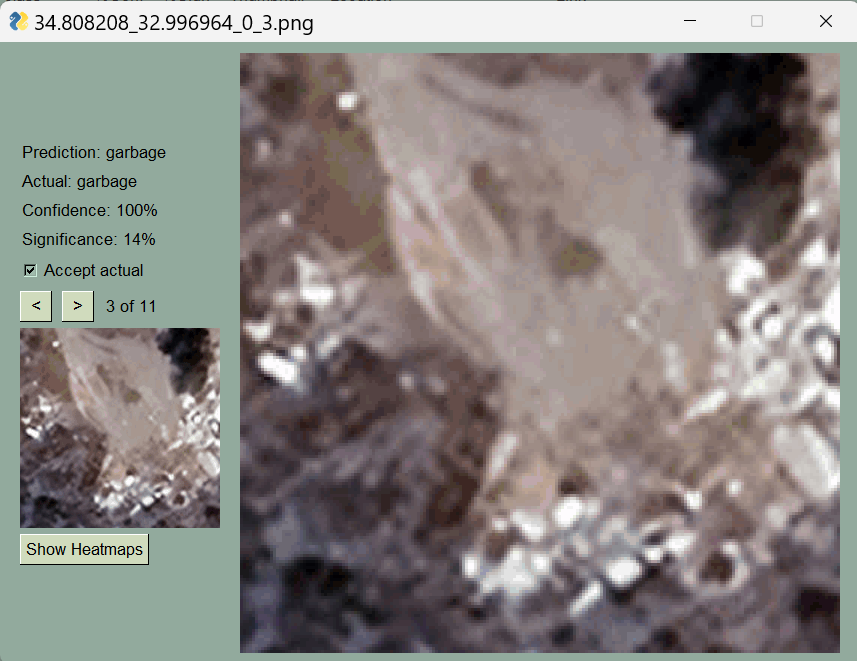}
\caption{File Details Popup}
\label{fig:file_details}
\end{figure}

\subsubsection{Image Details and Heatmap Visualization}
\begin{figure}[!h]
\centering
\includegraphics[width=\columnwidth]{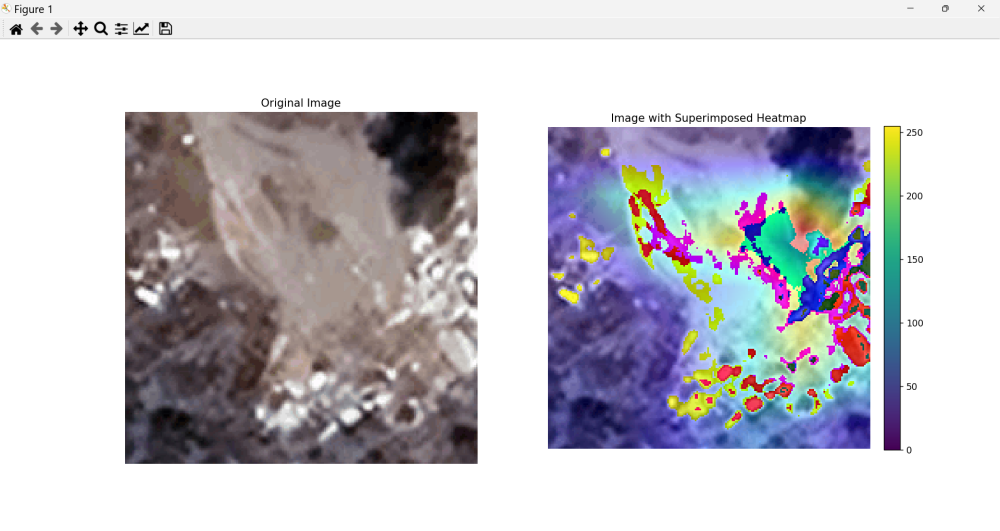}
\caption{Heatmap Popup}
\label{fig:file_heatmap}
\end{figure}

Each image thumbnail functions as a button that opens a file details window upon being clicked (see Figure \ref{fig:file_details}). This popup provides comprehensive information about the selected image and includes several interactive features:

\begin{itemize}[nosep]
    \item Prediction Toggle: Users can toggle the recorded prediction for the image, enabling adjustments when creating labeled datasets from a prediction run. This feature is particularly useful for refining datasets for training or testing purposes.
    \item Enlarged Image View: The popup displays an enlarged representation of the image, aiding users in visually confirming or assigning labels to images with greater accuracy.
    \item Heatmap Visualization: A ``Show Heatmap" button (see Figure \ref{fig:file_heatmap}) allows users to view a heatmap representation associated with the image. This visual explanation highlights the areas of the image that contributed most to the prediction, offering valuable insights into the model's decision-making process and helping diagnose potential issues in predictions.
    \item Geophysical Location: A ``Show on Google Maps" button is available to display the geographical location of the image patch, where applicable. This feature overlays the location on Google Maps, providing a visual reference for the origin of the image.
\end{itemize}

These features collectively enhance the usability and interpretability of the tool, empowering users to refine their datasets and better understand model behavior.

%% file: case_studies.tex
\section{Case Studies}

deepTerra has been applied to a variety of situations, including garbage detection, private swimming pool identification, beehive locality, and various others.  Full details of these case studies can be found in \cite{wilkinson_deepterra_nodate}.

\subsection{Garbage Dump Detection}
\begin{figure}[ht!]
    \centering
    \includegraphics[width=0.3\textwidth]{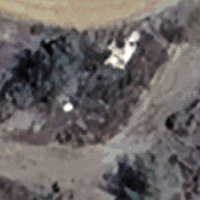}
    \caption{An image containing garbage.}
    \label{fig:garbage}
\end{figure}

Garbage dump recognition presents various challenges, including:

\begin{itemize}[nosep]
    \item High dimensionality of data
    \item Utilizing high-resolution satellite image sources
    \item Irregular, difficult-to-categorize garbage targets
    \item Obstructions by trees and vegetation
    \item Rapidly evolving heat characteristics
    \item Variation according to historical snapshots
\end{itemize}

deepTerra was used to help produce and organize a dataset of a few hundred verified and labeled garbage/not garbage images.  The image augmentation tool boosted this to a dataset of a few thousand.  A sample image is reproduced in figure \ref{fig:garbage}.  

The dataset was combined with the ResNet50 CNN architecture to train and evaluate a model that could recognize garbage in an independently gathered test dataset with 88\% accuracy.

The model has been applied to a region of Cyprus known as Pitsilia \cite{wikipedia_pitsilia_nodate}, a mountainous area covers an area of \SI{280}{\square\kilo\metre}. The system started at a north east latitude/longitude coordinate and automatically downloaded a set of high-resolution images from google earth that cover the entire region.  
The tool automatically organized these as a set of images patches suitable for input into the model.  This resulted in 362,880 images, each representing \SI{20}{\square\metre} ground area.  The tool also calculated the latitude/longitude coordinates for later use.

\begin{figure}[ht!]
    \centering
    \includegraphics[width=\columnwidth]{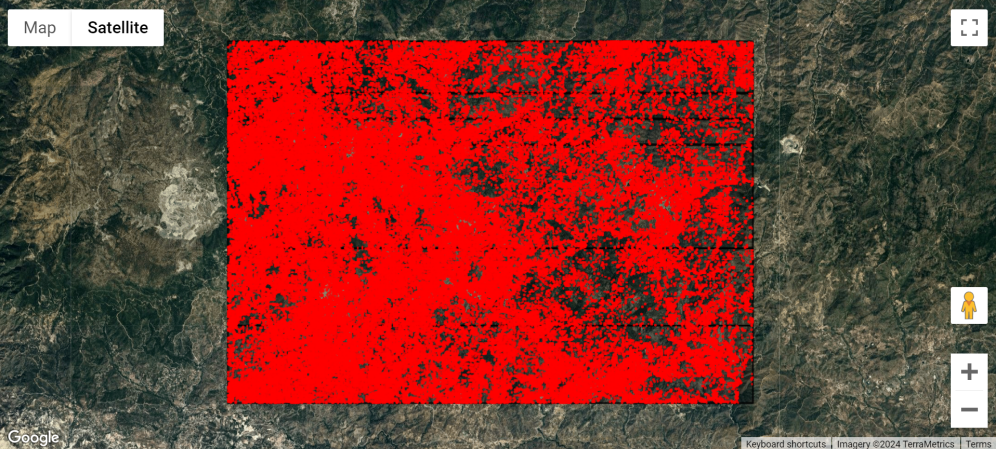}
    \caption{Google map overlay at low level of zoom.}
    \label{fig:pitsilia_big}
\end{figure}

The model was used to perform predictions against each image. Garbage was predicted to occur in 94,288 images patches, or 26\% of those surveyed. In addition, a zoomable Google maps overlay was generated showing each image patch location predicted to contain garbage.  This is illustrated in figure \ref{fig:pitsilia_big}.

Garbage detection in Cyprus using these processes was the basis of an MSc thesis \cite{wilkinson_identification_2023-1}.  Work is ongoing in this area, and updates are regularly posted \cite{wilkinson_deepterra_nodate}.

\newpage
\subsection{The Private Pools Project}

The main aim of this case study was to demonstrate the benefits of image augmentation.
The study focused on identifying private swimming pools using the VGG19 model, chosen for its simplicity and suitability for this task.  A model was trained on non-augmented and augmented datasets, and tested on a independently collected dataset. The non-augmented model gave an accuracy of 72\% whilst the augmented model one of 81\%.  This helps illustrate that image augmentation can improve accuracy of a model.

\subsection{Beehive Project}
\begin{figure}[ht!]
    \centering
    \includegraphics[width=0.3\textwidth]{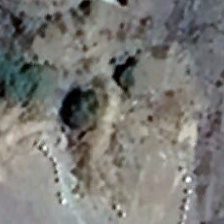}
    \caption{An image containing beehives.}
    \label{fig:beehives}
\end{figure}

Here the aim was to apply the tool to very small and difficult to collect datasets.  For this, the target objects to be detected were beehives.  A sample images is shown in Figure \ref{fig:beehives}

A small collected dataset of 24 images was used in conjunction with the Xception architecture. 
When applied to a test dataset of 60 images an accuracy of 84\% was obtained.  This illustrates that useful models can be developed even with very limited datasets.  One thing worth noting is that the turnaround time to collect and organize training and test images, train a few models, run the best performing one on the test dataset and obtain validation metrics, was around an hour from start to finish.

\subsection{Cats vs Dogs}

Whilst deepTerra has been designed to operate on satellite images, it can work equally well as an arbitrary image classifier.  To demonstrate this, the popular cats vs dogs dataset \cite{noauthor_kaggle_nodate} was used to train a model to distinguish between cats and dogs. A subset of 1000 images (500 cat, 500 dog) was used to train a ResNet50 model. When applied to a separate test dataset, this achieved an impressive accuracy of 98\%, failing to correctly categorize just 3 images from the 200 image test data set.

%% file: future_plans.tex
\section{Future Plans}
As technology evolves, deepTerra is committed to advancing its capabilities to address emerging needs and challenges. Future development efforts focus on integrating cutting-edge features and expanding the tool’s versatility across various domains. Below are some of the planned enhancements being explored:

\begin{itemize}[nosep]
\item Multicategory Classification: Expanding the tool to handle datasets with multiple classes, enabling classification tasks across a broader range of categories and use cases.
\item Advanced AI Techniques: Implementing innovative approaches such as hybrid models, which combine multiple architectures for enhanced performance, and incremental learning, allowing the system to adapt continuously as new data becomes available.
\item Multispectral Imaging: Incorporating multispectral data analysis to detect features like heat signatures and other spectral characteristics, broadening the range of applications.
\item Image Segmentation: Adding capabilities for image segmentation, enabling pixel-level classification and precise delineation of regions within images.
\item Historical Analysis: Introducing tools for analyzing historical datasets, allowing users to identify trends, monitor changes over time, and make informed predictions based on past data.
\item Real-time Processing for UAVs: Developing a real-time module optimized for mobile platforms, such as unmanned aerial vehicles (UAVs), to support on-the-fly image detection and analysis in dynamic environments.
\end{itemize}